\documentclass[letterpaper, 10 pt, conference]{ieeeconf} 
\IEEEoverridecommandlockouts
\usepackage{url}
\usepackage{float}
\usepackage{amsmath}
\usepackage{graphicx}
\usepackage{subcaption}
\usepackage{siunitx}
\usepackage{textcomp}
\usepackage{algorithm}
\usepackage{algorithmicx}
\usepackage{algpseudocode}
\usepackage[thinlines]{easytable}
\usepackage[colorlinks = true,
            linkcolor = blue,
            urlcolor  = blue,
            citecolor = blue,
            anchorcolor = blue]{hyperref}

\begin{document}
\title{Real-time Grasp Pose Estimation for Novel Objects in Densely Cluttered Environment}

% \title{Pick and Place of novel objects}
% \title{Grasping of Novel Objects for Robotic Pick and Place Applications}

% \author{ 
% Mohit Vohra, 
% Ravi Prakash,
% Laxmidhar Behera, \textit{Senior Member,~IEEE}
% }

\author{Mohit Vohra,
Ravi Prakash, and Laxmidhar Behera, \textit{Senior Member,~IEEE}
\thanks{All authors are with the Department of Electrical Engineering, Indian Institute of Technology Kanpur, Uttar Pradesh 208016, India. Email Ids: (mvohra, ravipr, lbehera)@iitk.ac.in.}%
}

\maketitle
\begin{abstract}
Grasping of novel objects in pick and place applications is a fundamental and challenging problem in robotics, specifically for complex-shaped objects. It is observed that the well-known strategies like \textit{i}) grasping from the centroid of object and \textit{ii}) grasping along the major axis of the object often fails for complex-shaped objects. In this paper, a real-time grasp pose estimation strategy for novel objects in robotic pick and place applications is proposed. The proposed technique estimates the object contour in the point cloud and predicts the grasp pose along with the object skeleton in the image plane. The technique is tested for the objects like ball container, hand weight, tennis ball and even for complex shape objects like blower (non-convex shape). It is observed that the proposed strategy performs very well for complex shaped objects and predicts the valid grasp configurations in comparison with the above strategies. The experimental validation of the proposed grasping technique is tested in two scenarios, when the objects are placed distinctly and when the objects are placed in dense clutter. A grasp accuracy of 88.16\% and 77.03\% respectively are reported. All the experiments are performed with a real UR10 robot manipulator along with WSG-50 two-finger gripper for grasping of objects.

% A novel real time grasp pose estimation technique for novel objects in robotic pick and place applications is proposed in this paper. The proposed technique estimates the object contour in point cloud and estimate the grasp pose along the object skeleton in the image plane. The technique is tested for the objects like ball container, hand weight, tennis ball and even for a complex shape objects like blower (non convex). Further the proposed technique is compared with the two other strategies \textit{i}) grasping from the centroid of object and \textit{ii}) grasping along the major axis of the object. It is observed that above two strategies fails for complex shaped objects while our method can predict the valid grasp configurations. The proposed technique is tested in two scenarios, when the objects are placed distinctly and when the objects are placed in a dense clutter. and overall we achieve a grasp accuracy of 88.16\% and 77.03\% respectively. All experiments are performed with UR10 robot manipulator along with WSG 50 two finger gripper for grasping of objects.
\end{abstract}

% \begin{IEEEkeywords}
% novel objects, grasping, object contour, object skeleton
% \end{IEEEkeywords}
\section{Introduction}
With the rapid growth in the e-commerce world, order volumes are increasing tremendously. To fulfill customer requirements, many industries and researchers have shifted their interest in warehouse automation. One of the main components of warehouse automation is the autonomous grasping of objects. Grasping of the object is a two-step procedure: perception and planning. The goal of the perception module is to estimate the grasp configurations either by estimating the pose of the object or by segmenting the object region and using some post-processing steps. The goal of the planning module is to generate the required trajectory and motor torques to achieve the desired configuration. Traditional grasping techniques evaluates the grasp configuration according to some metric \cite{ferrari1992planning} \cite{nguyen1988constructing} \cite{pokorny2013classical} on the assumption of simplified contact model, Coulomb friction and accurate 3D object model. But these methods often fail in the practical scenario due to inconsistencies in the simplified object and contact models.
\par
On the other hand, data-driven method \cite{bohg2014data} \cite{saxena2008robotic} has shown some significant results on the real data but requires an extensive training data set. In \cite{mahler2016dex}, the author developed a data set including millions of point cloud data to train a grasp quality CNN (GQ-CNN) with analytic metric and used GQ-CNN to select the best grasp plan which achieves 93\% success rate with eight types of known objects. \cite{pinto2016supersizing} requires $50$k grasping attempts to generate the training data set while \cite{levine2018learning} used $14$ robots to randomly grasp over $800,000$ times and collected grasping data to train CNN that teach robots to grasp. The main drawback of the data-driven methods is the poor performance on the novel object \textit{i.e.} objects which are not present in the data set, and in the warehouse industry because of the large variety of objects, it is difficult to include each one in the data set. Hence these methods are not reliable for grasping of novel objects.
\par
Various solutions have proposed in the literature for the grasping of novel objects. In \cite{gualtieri2018pick}, the author has formulated the grasping problem as a deep Reinforcement learning problem and has successfully demonstrated for unknown item with sampled grasp configurations. In \cite{gualtieri2016high} \cite{ten2017grasp}, author first samples the several thousand grasp candidates and then use CNN for selecting the best grasp configuration. Similarly, \cite{boularias2015learning} segments the object region in the image with the assumption that the object surface is convex, and select the best grasp configuration among the sampled grasp configuration. One of the main drawbacks of the above method is that the final grasp configuration will always be selected from the set of sampled grasp configurations. Hence there is no surety that the final grasp configuration is optimal. So to get the near optimal grasp configuration, thousands of samples need to be evaluated, hence requires massive computation.

\begin{figure}[t!]
\centering
  \includegraphics[width=7cm, height=6cm]{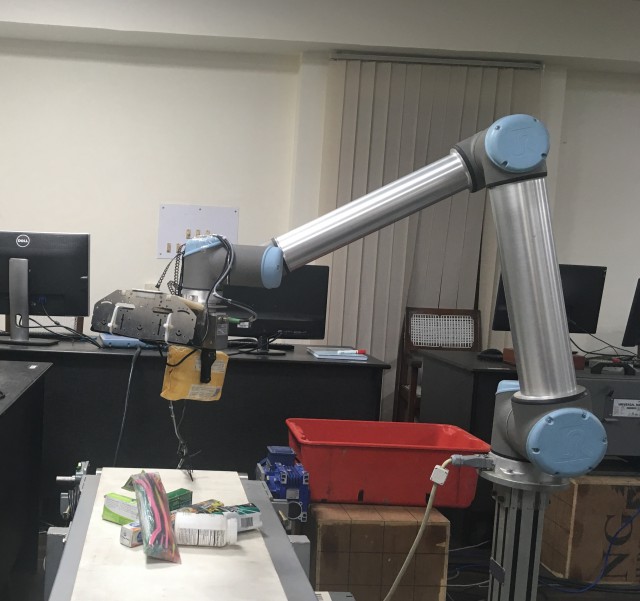}
  \caption{Grasping of the object}
  \label{fig:object_grasping}
  \vspace{-5mm}
\end{figure}

Various solutions have been proposed in the literature for directly predicting the grasp configurations (\textit{i.e.} no sampling is required). In \cite{katz2014perceiving}, authors develop a supervised learning framework with manually designed features as inputs, for predicting the affordances (push, pull or grasp) for each object, and selects the best affordance and its associated action to execute. \cite{tobin2018domain} trains a deep network on millions of synthetic images for predicting the grasp configuration. With the idea of domain randomization authors has demonstrated for a single and isolated object, that network trained on synthetic data can perform well on real data. In \cite{ten2016localizing}, the author develops an approach which can localize the handle-like grasp affordances in 3D point cloud data by satisfying the geometric constraints. \cite{vezzani2017grasping} model the pose computation problem as a nonlinearly constrained optimization problem with the object and gripper both are modeled as superquadric functions. 
\par
In this paper, we consider the problem of grasping of novel objects. Our framework consists of estimating the boundary points in the point cloud data, segmenting the object region in the image plane, estimate the skeleton the object region, followed by predicting the grasp configuration at each skeleton location and selection of best grasp configuration. The main advantage of our approach is that this approach is simple and can easily be adapted for other grippers as well. The second advantage is that our framework does not require any grasp configuration sampling step, which makes it more stable. Here stable means that for the same input, the developed framework will predict the same grasp configuration while sampling based method will produce different grasp configuration for each trial.
\par
The remainder of this paper is organized as follows. Gripper representation and problem formulation are presented in section II. A detailed explanation of our method is given in Section III. Experimental results for robotic grasping and quantitative results are mentioned in section IV.  The paper is finally concluded in Section V.

\section{Gripper representation and Problem Statement}

\subsection{Gripper representation}
\begin{figure}[!h]
 \centering\includegraphics[height=0.45\linewidth,width=0.5\textwidth]{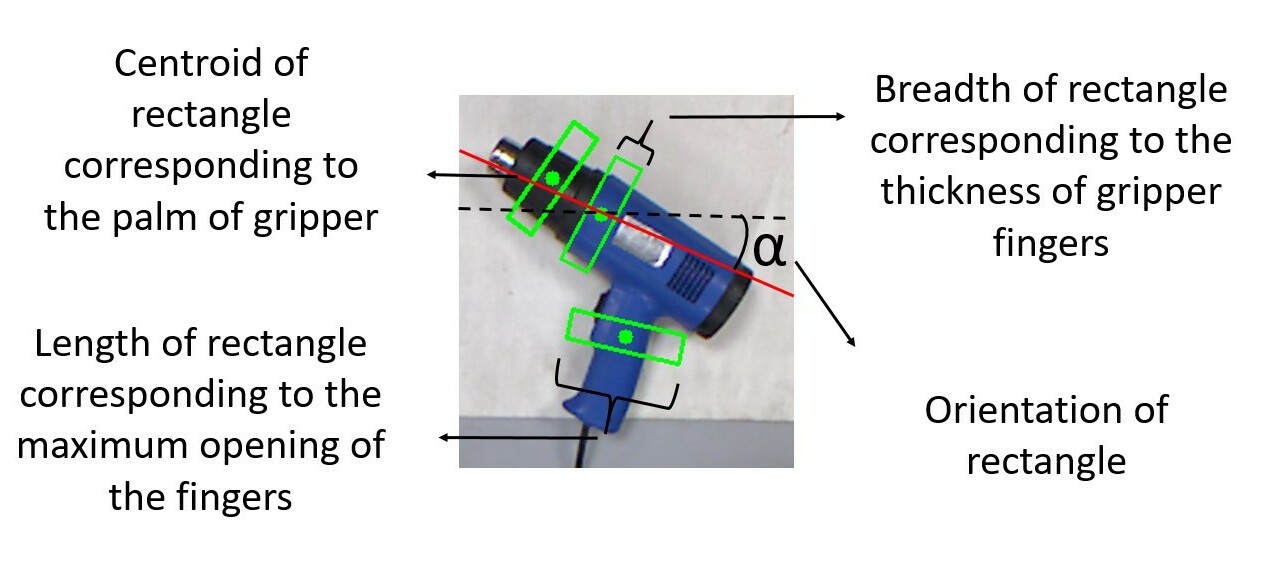}
\caption{Gripper representation}
\label{fig:gripper_rep}
\end{figure}

We represent two-finger gripper as a rectangle in the image plane, shown in Fig. \ref{fig:gripper_rep}. This representation of gripper is taken from \cite{jiang2011efficient} where length represents the maximum opening between the fingers, breadth represents the thickness of gripper fingers, the centroid of the rectangle is the palm of the gripper, and \boldmath$\alpha$ represents the orientation of gripper. Thus gripper configuration is represented by three parameters \textit{i.e.} $x$, $y$, \boldmath$\alpha$ where $x$, $y$ represents the centroid pixel location and \boldmath$\alpha$ represents the orientation.  

\subsection{Problem Statement}
Given the current image $\mathcal{I}$ and Point cloud $\mathcal{P}$ of the scene or workspace, find the gripper configuration ($x$, $y$, \boldmath$\alpha$) to grasp the objects present in the workspace. Main steps in our approach are to estimate the skeleton of the object and at each skeleton location predict the grasp configuration using local object skeleton structure and for selecting the final grasp configuration, point cloud data corresponding to the rectangular region (\textit{i.e.} gripper representation) will be used. A detailed explanation of each step is given in section III.
\section{Proposed Methodology}

\begin{figure}[!h]
 \centering
\includegraphics[height=1.5\linewidth, width=0.45\textwidth]{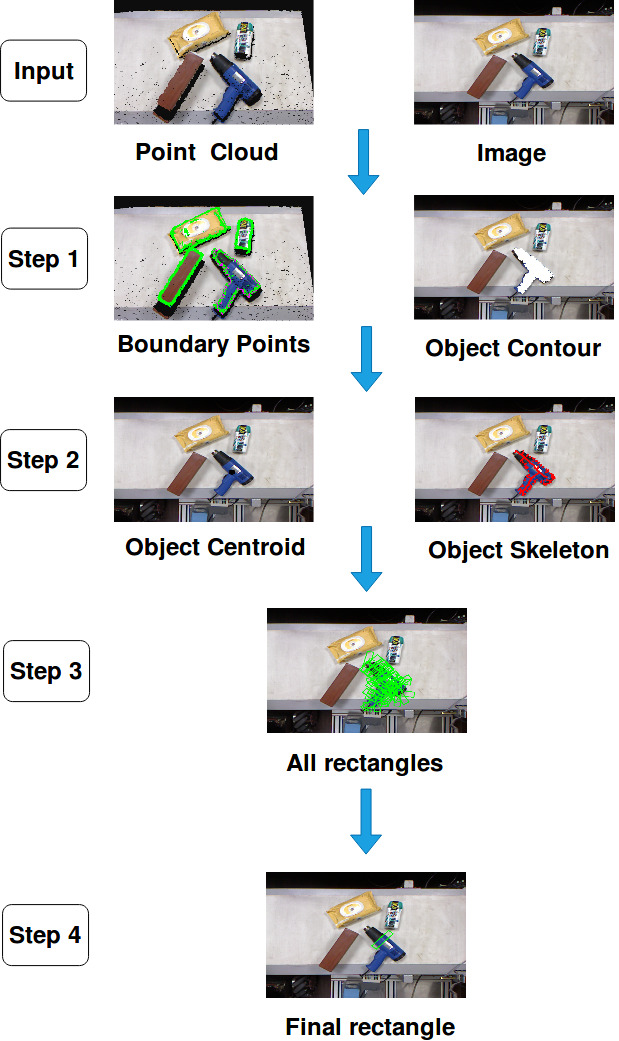}
\caption{Complete Pipeline}
\label{fig:complete_pipeline}
\end{figure}

The proposed method for estimating the grasp configuration is a four-step procedure. In the first step,  boundary points of the object are estimated using the point cloud data, and the corresponding object region is estimated in the image plane. In the second step, centroid and skeleton of the object are estimated. In the third step, grasp rectangles at each skeleton point are estimated, and in the final step, point cloud data corresponding to the grasp rectangle part and the centroid of the object is used to decide the final grasp rectangle or grasp configuration. Complete pipeline for estimating the final grasp configuration is shown in Fig. \ref{fig:complete_pipeline}.

\subsection{Boundary points and object contour}

\subsubsection{Boundary points in point cloud}
Let $\mathcal{P}$ be the raw point cloud data, $r_s$ represents the user defined radius and $t_h$ is user defined threshold value. Let $\mathcal{E}$ is the point cloud data which contains the boundary points. To decide if a given query point $\mathbf{p_{i}} \in \mathcal{P}$ is a boundary point or not, we calculate $\mathcal{R}(\mathbf{p_{i}})$, which is defined as set of all points inside a sphere, centered at point $\mathbf{p_{i}}$ with radius $r_s$. In point cloud, this is achieved through a k-dimensional (K-d) tree. We call each point in set $\mathcal{R}(\mathbf{p_{i}})$ as neighboring point of $\mathbf{p_{i}}$ and it can be represented as $\mathcal{R}(\mathbf{p_{i}}) = \begin{Bmatrix}
\mathbf{n_{1}, n_{2}, \dots , n_{k}}
\end{Bmatrix}$. For each query point $\mathbf{p_{i}}$ and neighboring point $\mathbf{n_{j}}$ we calculate the directional vector as

\vspace{-5mm}

\begin{align}
% \label{eqn:eq1}
 \mathbf{d(p_{i}, n_{j}) = n_{j} - p_{i}}\\
 \mathbf{\hat{d}(p_{i}, n_{j})} = \frac{\mathbf{d(p_{i}, n_{j})}}{\left \| \mathbf{d(p_{i}, n_{j})} \right \|}
\end{align}

Then we calculate the resultant directional vector $\mathbf{\hat{R}(p_{i})}$ as sum of all directional vector and normalize it
\begin{align}
% \label{eqn:eq2}
 \mathbf{R(p_{i})} = \sum_{j=1}^{k}\mathbf{\hat{d}(p_{i}, n_{j})},
 \mathbf{\hat{R}(p_{i})} = \frac{\mathbf{R(p_{i})}}{\left \|  \mathbf{R(p_{i})}  \right \|}
\end{align}

We assign a score $s(\mathbf{p_{i}})$ to each query point $\mathbf{p_{i}}$  as average of dot product between $\mathbf{\hat{R}(p_{i})}$ and $\mathbf{\hat{d}(p_{i}, n_{j})}$ for all neighboring points.
\begin{align}
% \label{eqn:eq3}
 s(\mathbf{p_{i}}) = \frac{\sum_{j=1}^{k}\mathbf{\hat{R}(p_{i})} \cdot \mathbf{\hat{d}(p_{i}, n_{j})}}{k}
\end{align}

If $s(\mathbf{p_{i}})$ is greater than some threshold $t_h$ then $\mathbf{p_{i}}$ is considered as a boundary point else not. The boundary points in point cloud data is shown in Fig. \ref{fig:complete_pipeline}, step 1 output.

\subsubsection{Close contour in point cloud}
Once we got the boundary points, the next step is to select the boundary points, which forms a closed loop. This step ensures that unwanted boundary points because of sensor noise or because of any sharp features on the surface of the object can be ignored, thus avoiding the unnecessary computation. Following are the steps to find the closed loop points:

\begin{itemize}
    \item Select the the starting point $\mathbf{s_i}$ with lowest \textit{z} value (closest to Kinect).
    \item Let the nearest neighbour of $\mathbf{s_i}$ be $\mathbf{n_j}$.
    \item Remove $\mathbf{s_i}$ from boundary point set and store $\mathbf{s_i}$ in an array $\mathbf{A}$.
    \item Find the nearest neighbour of $\mathbf{n_j}$, since we have removed $\mathbf{s_i}$, so we will get the nearest neighbor of $\mathbf{n_j}$ different from $\mathbf{s_i}$.
    \item Store $\mathbf{n_j}$ in $\mathbf{A}$ and repeat the step $3$ and $4$ with $\mathbf{n_j}$ till
    \begin{itemize}
        \item Nearest neighbor is same as $\mathbf{s_i}$, hence all points in $\mathbf{A}$ forms a close loop.
        \item If there is no neighbor, then remove last element from $\mathbf{A}$, find the neighbor for last element of $\mathbf{A}$ and repeat the step $3$ and $4$.
    \end{itemize}
\end{itemize}

The boundary points which forms a closed loop are shown in pink color (Fig. \ref{fig:complete_pipeline}, step 1 output)

\subsubsection{Object region in image plane}
Since Kinect gives the registered point cloud data, so we can find the pixel locations corresponding to the closed loop boundary points and using the standard flood fill algorithm we will find the object region in the image plane which is shown in Fig. \ref{fig:complete_pipeline}, step 1 output. 

\subsection{Centroid and skeleton of object in image plane}
\subsubsection{Object Centroid}
Output of a flood fill algorithm is a binary mask with white pixels represents the object region and black pixels represents the non object region. We define centroid of the object as the average of the object pixel locations which is shown in Fig. \ref{fig:complete_pipeline}, step 2 output.

\subsubsection{Skeleton of object}
In this work, we mainly focus on grasping the objects vertically using two-finger gripper. Hence we assume that grasping along the central axis of the object surface, has a high probability of successful grasping as compared to randomly selecting the grasp configurations. To find the central axis of the object surface, we find the skeleton of the object region in the image plane. Skeletonization reduces the foreground regions to a skeletal remnant that largely preserves the shape and connectivity of the original regions. Since the output of skeletonization is pixel locations which are along the central axis of the object region, hence this step provides several pixel locations for calculating the grasp configurations, instead of only the centroid pixel. Result of skeletonization is shown in Fig. \ref{fig:complete_pipeline}, step 2 output.

\subsection{Grasp rectangles at each skeleton point}
Since a gripper is represented by three parameters (center of the rectangle, $x$, $y$, and orientation of the rectangle \boldmath$\alpha$), here skeleton pixel represents the center  and to find the orientation, we fit the straight line on the local skeleton structure around the skeleton pixel using orthogonal linear regression\footnote{https://www.codeproject.com/Articles/576228/Line-Fitting-in-Images-Using-Orthogonal-Linear-Reg}. 
\par
\begin{figure}
\includegraphics[scale=0.25]{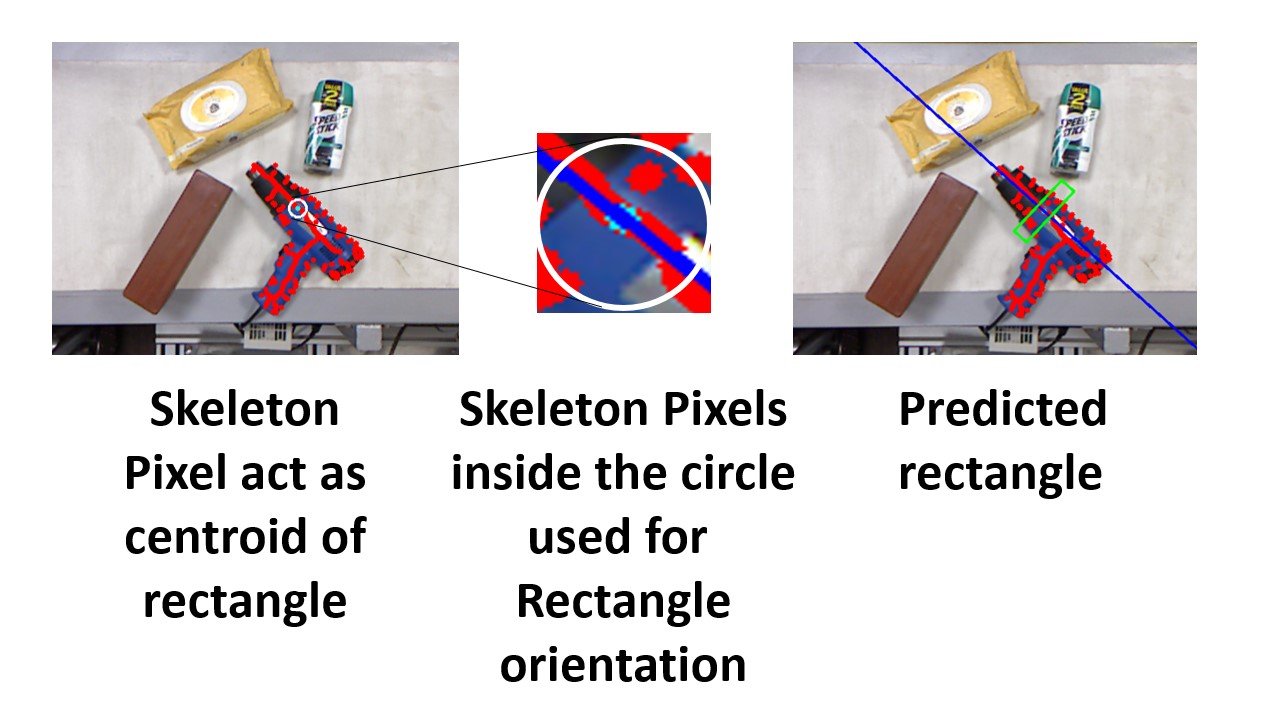}
\caption{Grasp Rectangle}
\label{fig:grasp_rectangle}
\end{figure}

 Let $q_j$ represents the skeleton pixel location where we want to estimate the grasp configuration. For visualization, a white circle is drawn at a pixel location of $q_j$, which is shown in Fig. \ref{fig:grasp_rectangle}. All skeleton pixels inside the circular region is local skeleton structure around $q_j$. To find the slope of line let us assume that $p_i$ represents the $i^{th}$ skeleton pixel inside the circular region, $p_i(r)$, $p_i(c)$ represents the row and column number of $p_i$ respectively and $n$ be the total number of pixels inside the circular region.
\begin{align}
 \mu_r = \frac{\sum_{i=1}^{n}p_{i}(r)}{n}
 \end{align}
 \vspace{-4mm}
 \begin{align}
 \mu_c = \frac{\sum_{i=1}^{n}p_{i}(c)}{n}
 \end{align}
 \vspace{-4mm}
 \begin{align}
\sigma_r = \sum_{i=1}^{n}(\mu_r - p_{i}(r))^2
 \end{align}
 \vspace{-4mm}
 \begin{align}
\sigma_c = \sum_{i=1}^{n}(\mu_c - p_{i}(c))^2
 \end{align}
 \vspace{-4mm}
 \begin{align}
\sigma_{rc} = \sum_{i=1}^{n}(\mu_r - p_{i}(r))(\mu_c - p_{i}(c))
 \end{align}
  \vspace{-4mm}
 \begin{align}
b =\frac{ \sigma_r - \sigma_c + \sqrt{ ( \sigma_r - \sigma_c)^2 + 4 \sigma_{rc}^2}}{2 \sigma_{rc}}
\end{align}
 \vspace{-4mm}
 \begin{align}
slope = \tan^{-1}(b)
\end{align}
\vspace{-5mm}

For visualization, we have shown the line with calculated slope passing through the $q_j$ and grasp rectangle at $q_j$ with an orientation of rectangle is same as the slope of the line which is shown in Fig. \ref{fig:grasp_rectangle}.

\subsection{Final grasp selection}
\begin{figure}[!h]
\centering
  \begin{subfigure}[b]{0.45\linewidth}
    \includegraphics[width=\linewidth]{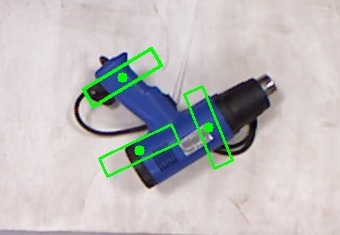}
    \caption{Image}
  \end{subfigure}
    \begin{subfigure}[b]{0.48\linewidth}
    \includegraphics[width=\linewidth]{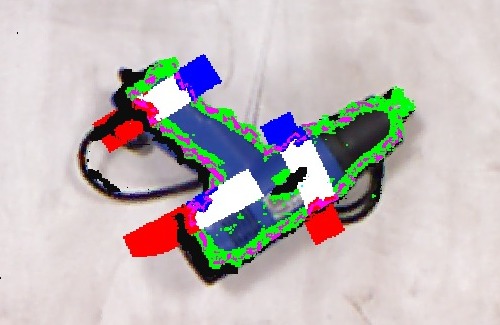}
    \caption{Point Cloud}
  \end{subfigure}
  \caption{Grasp configurations}
  \label{fig:Grasp configurations}
  \vspace{-4mm}
\end{figure}

Once we have the number of possible grasp configurations in the image plane, as shown in Fig. \ref{fig:complete_pipeline}, step 3 output, next step is to select the final grasp configuration. For each grasp configuration, we partition the point cloud data into three regions \textit{i.e.} object region (white) and two non object regions (red and blue) as shown in Fig. \ref{fig:Grasp configurations}. We consider two conditions for configuration to be valid
\begin{itemize}
    \item Object region must always exist in between the two non object regions.
    \item Difference between the \textit{z} values of each point in non object region and \textit{z} value at centroid should be greater than some threshold value \textit{i.e.} $z(p_i) - z(centroid) > th$, where $p_i$ is any point in the non object region. This threshold mainly depends on the length of the gripper's finger.
\end{itemize}

Above conditions filter out the grasp configurations at the boundary of objects which has a high chance of gripper finger collisions with other objects. From the filtered configurations, we select the grasp configuration which is closest to the object centroid by calculating the distance between the object centroid and grasp rectangle centroid.

\section{Experiments and Result}

\begin{figure}[t!]

  \includegraphics[width=9cm, height=5cm]{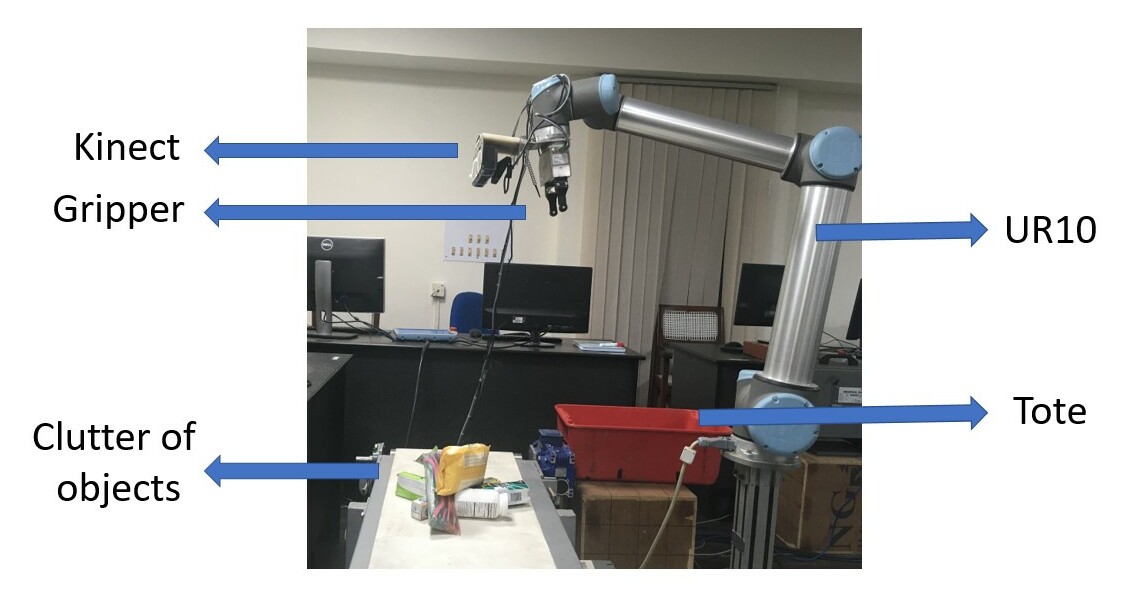}
  \caption{Hardware Setup}
  \label{fig:hardware_setup}
  \vspace{-5mm}
\end{figure}

\subsection{Experimental Setup}
Our robot platform setup shown in Fig. \ref{fig:hardware_setup} consists of a UR10 robot manipulator with its controller box (internal computer) and a host PC (external computer). The UR10 robot manipulator is a 6 DOF robot arm designed to safely work alongside and in collaboration with a human. This arm can follow position commands like a traditional industrial robot, as well as take velocity commands to apply a given velocity in/around a specified axis. The low level robot controller is a program running on UR10's internal computer broadcasting robot arm data, receiving and interpreting the commands and controlling the arm accordingly. There are several options for communicating with the robot low level controller to control the robot including the teach pendent or opening a TCP socket (C++/Python) on a host computer. We used eye-in-hand approach i.e. the vision hardware consisting of RGB-D Microsoft Kinect sensor mounted on the wrist of the manipulator. The WSG 50 robotic parallel gripper made by SCHUNK, is used to grasp the objects. ROS drivers are used to communicate between the sensor, manipulator and the gripper.

\begin{figure*}
 \begin{subfigure}[b]{0.2\textwidth}
                \centering
                \includegraphics[scale=0.12]{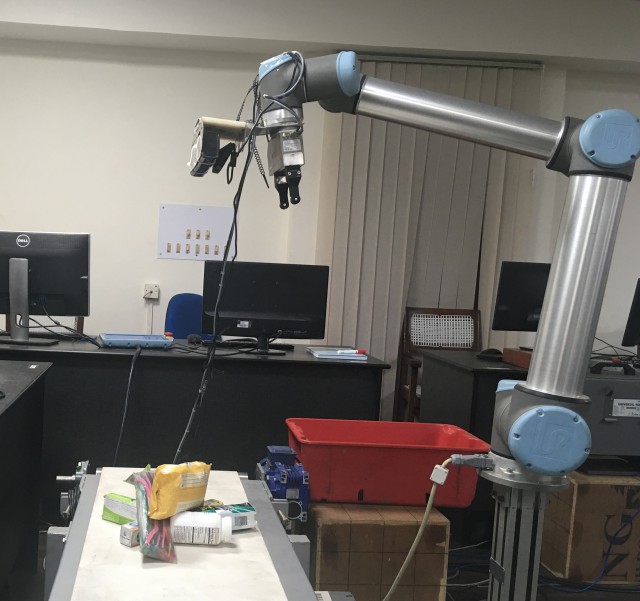}
                \caption{Calling vision service}
                \label{fig:gull}
        \end{subfigure}%
        \begin{subfigure}[b]{0.2\textwidth}
                \centering
                \includegraphics[scale=0.12]{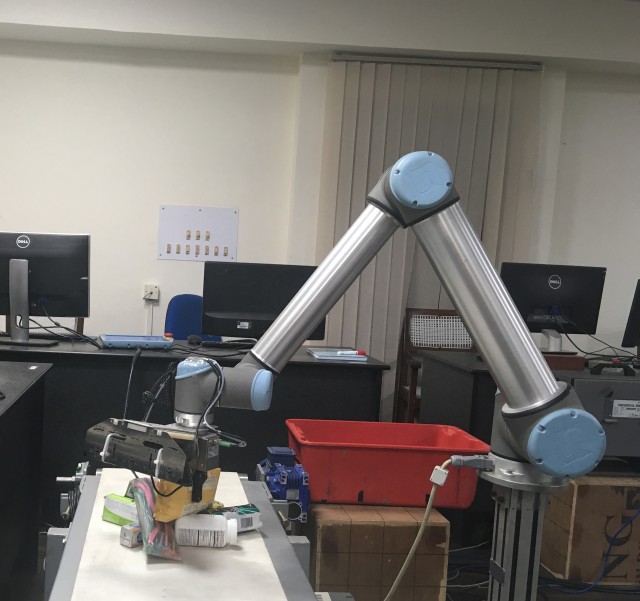}
                \caption{Grasp Object}
                \label{fig:gull}
        \end{subfigure}%
        \begin{subfigure}[b]{0.2\textwidth}
                \centering
                \includegraphics[scale=0.12]{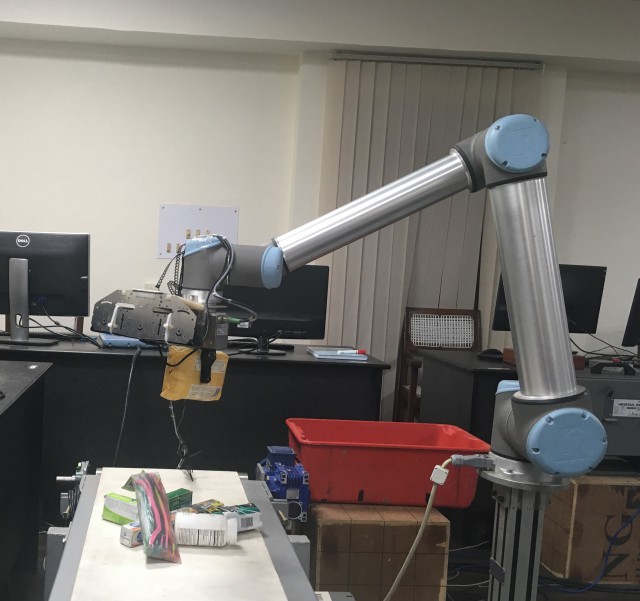}
                \caption{Lift object}
                \label{fig:gull2}
        \end{subfigure}%
        \begin{subfigure}[b]{0.2\textwidth}
                \centering
                \includegraphics[scale=0.12]{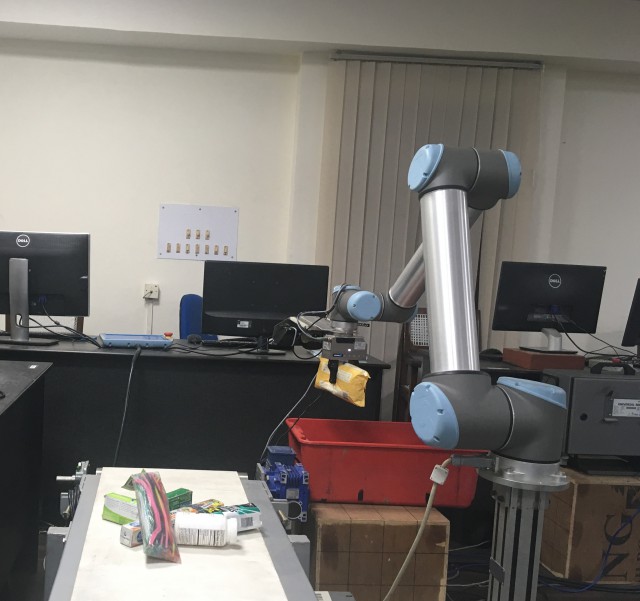}
                \caption{Place object}
                \label{fig:tiger}
        \end{subfigure}%
                \begin{subfigure}[b]{0.2\textwidth}
                \centering
                \includegraphics[scale=0.12]{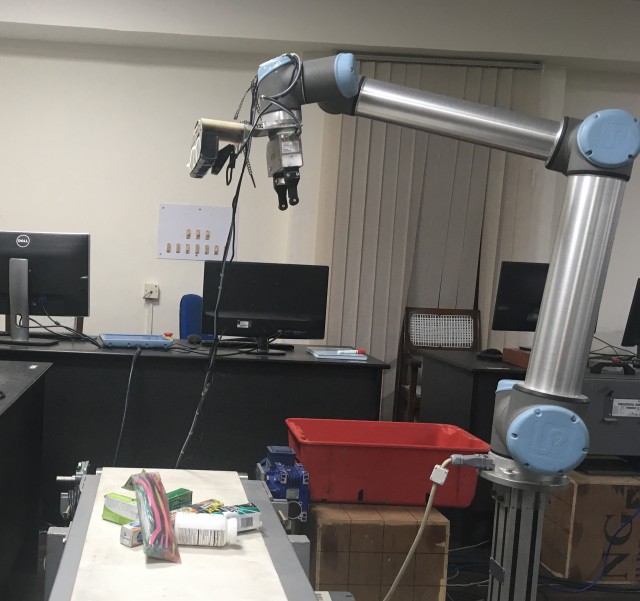}
                \caption{View Pose}
                \label{fig:mouse}
        \end{subfigure}
        \caption{Execution steps during robotic manipulation}\label{fig:full_system}
        \vspace{-4mm}
\end{figure*}

\subsection{Motion Control Module}
Order in which the the actions are executed to grasp the objects is shown in Fig. \ref{fig:full_system}. 

\textbf{\textit{i) Calling vision service:}} In first step vision service estimates the grasp configuration ($x$, $y$, \boldmath$\alpha$). Here $x$, $y$, are the pixel coordinates, so from the registered point cloud data we can get the 3D location and since we always grasp the object vertically, hence \boldmath$\alpha$ represents the angle of rotation around the gripper axis. Hence from $x$, $y$, \boldmath$\alpha$, we calculate the final pose of the end effector.

\textbf{\textit{ii) Grasp Object:}} This task defines the trajectory from current pose to the final grasp pose estimated from vision service. Since final pose will always exist at the object surface, so to grasp the object vertically we define one intermediate trajectory point which is at some fixed height and use ROS modules to generate trajectory that starts at current pose, passes through intermediate point and ends at final pose. After execution of the trajectory gripper close command is triggered to grasp the object.  

\textbf{\textit{iii) Lift object:}} Once the object is grasped, we lift the object vertically up at some fixed height. 

\textbf{\textit{iv) Place object:}} Next task is to place the objects in a tote. In our experiments we have fixed the tote and fixed joint angle motion is executed to reach the tote. After execution of the joint angle motion, gripper open command is triggered to place the object.  

\textbf{\textit{v) View Pose:}} Final task is to reach at a desired joint space configuration so that the objects are in the proper view of the vision sensor. This motion is also the fixed joint angle motion and after the execution of the joint angle motion, vision service will be called and entire process will repeat till we clear the clutter.
 Video link for the real experiment is \href{https://www.youtube.com/watch?v=SziHDyDDUNU}{video link}.

\subsection{Objects Used for grasping}

\begin{figure}[t!]
\centering
  \includegraphics[width=7cm, height=4.5cm]{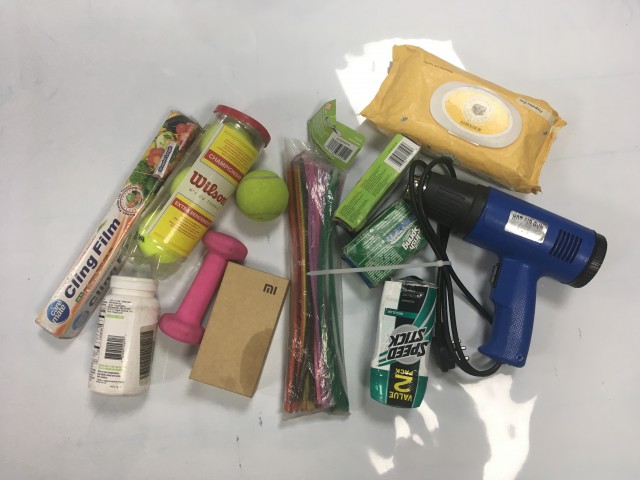}
  \caption{Objects used for experiments}
  \label{fig:objects}
  \vspace{-5mm}
\end{figure}

We have used $15$ objects for our experiments which is shown in Fig. \ref{fig:objects}. Most of the objects are very common. Objects are chosen in such a way that the thickness of object part should not be greater than the maximum opening of the gripper. We deliberately included the blower (blue gun like object) to test our algorithm for a complex shape object. For our experiments we randomly select the objects from the object set and place the objects either separately or in a clutter.

\subsection{Parameters setting}

 \textbf{\textit{i) Boundary points:}} Based on several trials we found that for extracting boundary points in point cloud data, $r_s$= 0.02m and $t_h$ = 0.35 are optimal values. It is because if we set $r_s$ at a minimal value, then the quality of the boundary points will be inferior and if we set $r_s$ at huge value than quality of boundary points will be very good but computation time will be very high. Similarly, if we set $t_h$ very low, then the quality of boundary points is inferior because almost every point in the point cloud data will be considered as a boundary point. If we set $t_h$ very high, then the only small number of point cloud data will be considered as boundary points, and hence there be less probability of finding the closed loop boundary points.

 \textbf{\textit{ii) Object region in image plane:}} To find the object region or mask in the image plane, we use the flood fill algorithm with a starting point or seed point as the average of pixels corresponding to the closed loop boundary points. Because of the complex shape of objects, there is a high chance that seed point can exist outside the actual object region. So to get the proper object region, we count the number of pixels corresponding to the object region in the mask, and if the number of pixels is more than the half of the image size, then we invert the mask else remains the same.

 \textbf{\textit{iii) Object Skeleton:}} To find the object skeleton, we use the standard morphological operations defined in OpenCV library \textit{i.e.} dilation and erosion with default parameter values.

 \textbf{\textit{v) Grasp Rectangle:}} Dimension of the rectangle depends on the maximum opening of the gripper and the thickness of the gripper finger. For WSG-50 gripper, the maximum opening is of 10cm, and the thickness of the finger is 2.5cm. In our experiments, during vision service, Kinect is at the height of 0.8m from the flat surface where objects will be placed, so we draw a rectangle of 10cm by 2.5cm on the flat surface and find its projection in the image plane which gives a rough dimension of the grasp rectangle. In our experiments, we set length of the rectangle at $80$ pixels and width at $20$ pixels.
 
  \textbf{\textit{v) Grasp Selection:}} We define a grasp configuration valid if difference between the \textit{z} values of each point in non object region and \textit{z} value at centroid is greater than some threshold value \textit{i.e.} $z(p_i) - z(centroid) > th$, where $p_i$ is any point in the non object region and we set $th$ to be 3cm.

 \subsection{Results and comparison}
 \subsubsection{Results}
 To test our method, we perform 25 rounds of experiments with 5 rounds when objects are placed separately, and in 20 rounds, we place objects in clutter. In the case of isolated objects, our method performs very well for larger and thicker objects. We believe that this limitation is because of the Kinect sensor noise. So if we use Kinect V2 or Ensenso, which produces high-resolution point cloud data then our method can estimate the grasp regions for thinner objects as well. Overall we got a grasping accuracy of 88.16\% when objects are placed separately, and in case of clutter, grasp accuracy drops to 77.03\%. This is because most of the objects are occluded or central part of the object is covered with other objects. Hence our method always predicts the grasp configuration near the corners of the object and decreases the grasp success probability.
 
 \subsubsection{Comparison}
\begin{figure}
\centering
  \begin{subfigure}[b]{0.3\linewidth}
    \includegraphics[width=1\textwidth]{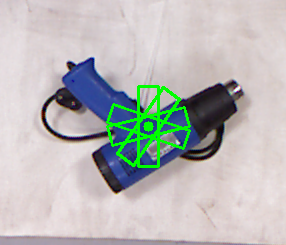}
    \caption{Strategy 1}
  \end{subfigure}
  \begin{subfigure}[b]{0.3\linewidth}
    \includegraphics[width=1\textwidth]{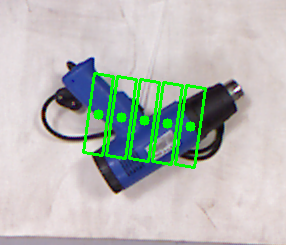}
    \caption{Strategy 2}
  \end{subfigure}
  \begin{subfigure}[b]{0.3\linewidth}
    \includegraphics[width=1\textwidth]{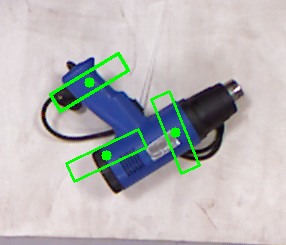}
    \caption{Proposd strategy}
  \end{subfigure}
  
  \caption{Grasp configurations for different strategies}
  \label{fig:Grasp comparison}
  \vspace{-4mm}
\end{figure}

 We compared the proposed method with two other strategies. In the first strategy, we grasp the objects from the centroid of the object region. In this strategy, each possible grasp configurations has the same centroid but different orientation.
 In the second strategy, we compute two orthogonal axes for the object region using PCA and generate the grasp configurations along the major axis, and each possible grasp configurations has the same orientation but different centroid. We found that for simple shaped objects when placed separately, all three strategies provides equivalent results \textit{i.e.} final grasp configuration is similar. But for complex shape objects, both strategy fails, while our strategy predicts some valid grasp configurations, as shown in Fig. \ref{fig:Grasp comparison}. So we conclude that the grasping from centroid or grasping along the major axis are special cases of our method, and our method is more general and can be applied for various shaped objects.

\section{Conclusion}
% A novel real time grasp pose estimation technique is proposed in this paper. Proposed technique extracts the object contour using point cloud data and estimate the grasp pose along the object skeleton estimated in the image plane. The technique is tested for the objects like ball container, hand weight, tennis ball and even for a complex shape objects like blower (blue gun like object). Proposed technique is compared with two other strategies \textit{i}) grasp from the centroid of object and \textit{ii}) grasp along the major axis of object, and it is observed that above two strategies fails for complex shaped objects while our method can predict the valid grasp configurations. The proposed technique is used for grasping of objects in two scenarios, when the objects are placed distinctly and when the objects are placed in a dense clutter, and overall we achieve a grasp accuracy of 88.16\% and 77.03\% respectively. All experiments are performed with UR10 robot manipulator along with WSG 50 two finger gripper for grasping the objects.

A novel grasp pose estimation technique was proposed and tested on various objects like a ball, hand weight, tennis ball container, and for complex shaped objects like blower (nonconvex shape). It is observed our method performs better than the well-known strategies \textit{i}) grasp from the centroid of object and \textit{ii}) grasp along the major axis of the object. Further, we tested the technique in two scenarios, when the objects are placed distinctly and when the objects are placed in a dense clutter, and overall we achieve a grasping accuracy of 88.16\% and 77.03\% respectively. All experiments are performed with UR10 robot manipulator along with WSG-50 two-finger gripper for grasping of objects.

We believe that if we could use a color-based segmentation or machine learning based semantic segmentation technique, then we can get the object contour and object skeleton with greater accuracy and can get better grasping.

\bibliography{citations} 
\bibliographystyle{ieeetr}
\end{document}